\pdfoutput=1

\documentclass[11pt]{article}

\usepackage[final]{acl}

\usepackage{txfonts}
\usepackage{latexsym}
\usepackage{array}

\usepackage[T1]{fontenc}

\usepackage[utf8]{inputenc}

\usepackage{microtype}

\usepackage{inconsolata}

\usepackage{graphicx}
\usepackage{multirow}

\title{Historical Ink: Semantic Shift Detection for 19th Century Spanish}

\author{
  Tony Montes$^1$ \quad Laura Manrique-Gómez$^2$ \quad Rubén Manrique$^1$  \\
  $^1$ Systems and Computing Engineering Department, Universidad de los Andes \\
  $^2$ History and Geography Department, Universidad de los Andes \\
  Bogotá D.C. \\
  \texttt{\{t.montes, l.manriqueg, rf.manrique\}@uniandes.edu.co} \\
}

\begin{document}
\maketitle
\begin{abstract}
This paper explores the evolution of word meanings in 19th-century Spanish texts, with an emphasis on Latin American Spanish, using computational linguistics techniques. It addresses the Semantic Shift Detection (SSD) task, which is crucial for understanding linguistic evolution, particularly in historical contexts. The study focuses on analyzing a set of Spanish target words. To achieve this, a 19th-century Spanish corpus is constructed, and a customizable pipeline for SSD tasks is developed. This pipeline helps find the senses of a word and measure their semantic change between two corpora using fine-tuned BERT-like models with old Spanish texts for both Latin American and general Spanish cases. The results provide valuable insights into the cultural and societal shifts reflected in language changes over time\footnote{The pipeline and code can be found at \url{https://github.com/historicalink/SSD-Old-Spanish}}.
\end{abstract}

\section{Introduction}

The study of how word meanings evolve over time, influenced by social, historical, and political factors, is a fundamental pursuit within linguistics and natural language processing. This evolution poses challenges in detection and interpretation, often addressed through the Semantic Shift Detection (SSD) task, also known as Lexical Semantic Change Detection task (LSCD) \cite{SSD, Hu2021}. Traditionally reliant on manual methods such as discourse analysis, recent computational linguistics advancements have revolutionized this field. These approaches streamline analysis and open doors to interdisciplinary research applications spanning sociology, history, and beyond, offering invaluable insights into cultural and societal shifts using digitized corpora.

In 2013, static word embeddings, also known as word vector representations, were first introduced by \citet{Mikolov} using the bag-of-words and skip-gram architectures. These embeddings represent words as static vectors that remain unchanged and are based on their surrounding words. \citet{Hamilton} first proposed using these embeddings for the SSD task by employing diachronic word2vec static embeddings to measure word meaning changes across consecutive decades. Various approaches have been explored to automate this task effectively. \citet{SSD} proposed using contextual embeddings instead to capture multiple meanings assigned to the same word due to polysemy and homonymy, which static embeddings cannot achieve. This was accomplished by comparing multiple BERT-like Language Models \cite{BERT} such as XLM-RoBERTa.

In this paper, we focus on two things: crafting a 19th-century Spanish corpus ($C_{old}$) from sources spanning 1800 to 1914 and creating a customizable pipeline for assessing the SSD task. Utilizing this pipeline, we analyze the semantic changes of a set of target words, for both the global context and the specific Latin-American context. We explore a variety of known and novel solutions for the SSD task by comparing the 19th-century Spanish corpus with the Spanish portion of the "EUBookShop" corpus as the modern corpus ($C_{new}$) \cite{large_spanish_corpus}\footnote{This portion was taken from the large Spanish corpus available at \url{https://huggingface.co/datasets/josecannete/large_spanish_corpus}}.

\section{Related Work}

Recent advances in Semantic Shift Detection have leveraged many computational approaches based on natural language processing techniques. Contextual embeddings, capable of capturing multiple-word usages and meanings, have been used in most of the state-of-the-art solutions, summarized by \citet{SSD} who defines a classification framework based on three dimensions of analysis: meaning representation (\textit{form-} and \textit{sense-oriented} approaches), time awareness (\textit{time-oblivious} and \textit{-aware}) and learning modality (\textit{supervised} and \textit{unsupervised}, referencing to the injection of external knowledge support like a dictionary), useful for \textit{Contextualized Semantic Shif Detection}.

\citet{wu_explore} and \citet{wu_use} explore transformer-based BERT models for detecting semantic change. Martinc et al. uses contextualized embeddings to capture shifts in word usage over time, outperforming traditional techniques like Word2Vec and Glove by leveraging BERT's dynamic word representations. \citet{wu_use} adopt an unsupervised approach, obtaining and clustering word representations to measure change over time, aligning with human judgments. Both studies underscore the effectiveness of BERT-based models in identifying and analyzing diachronic linguistic changes.

Although most of the research in the field of semantic change has been done on a wide scope of languages, Spanish hasn't played such an important role in this field, except for some research, like LSCDiscovery in Spanish, a task presented by \citet{LSCD}. This task has facilitated the development and evaluation of SSD systems in this language, accompanied by an unannotated Spanish corpus for both modern and old texts, which has a size of $22M$ and $13M$ tokens respectively. Additionally, the task paper highlighted effective techniques and approaches within the solutions. The most successful solution for the LCSDiscovery task was GlossReader, developed by  \citet{glossreader}, which involved fine-tuning XLM-RoBERTa, a Language Model trained on more than 100 languages, with old English datasets and employing the model zero-shot cross-lingual transferability of the model to build contextualized embeddings for Spanish, and using this fine-tuned model for SSD tasks. This approach has demonstrated good performance, especially in avoiding issues associated with word form bias and labor-intensive annotation requirements. These advancements underscore the increasing significance and potential of computational methodologies in enhancing our comprehension and automation of semantic shifts in multiple languages.

Also, \citet{Hu2021} present a set of methodological considerations for low-resource languages such as 15th-century Spanish, where a lower amount of data is available, and the data is not as clean as in other high-resource languages such as English and Mandarin Chinese, stating that common SSD techniques are also useful for these cases, but must be used carefully, under a set of considerations.

\section{Data}

Selecting the data is a crucial step for the reliability of the results. The LSCDiscovery shared task provides a useful corpus for old Spanish texts within the years 1810-1906, with a size of $13M$ tokens \cite{LSCD}. However, this paper aims to construct a larger old Spanish corpus, also adding more presence from Latin-American countries. The main sources selected and filtered for this corpus were \textbf{Project Gutenberg}\footnote{Available at \url{https://www.gutenberg.org/browse/languages/es}} which was filtered by language and by the given date ranges (1800-1914), \textbf{The British Library books}\footnote{Available at \url{https://huggingface.co/datasets/TheBritishLibrary/blbooks}} (portion from 1800-1899) which was also filtered by language \cite{blbooks}, and the \textbf{LatamXIX}\footnote{Available at \url{https://huggingface.co/datasets/Flaglab/latam-xix}} dataset from the \textit{Historical Ink} project which contains Latin American texts from newspapers within years 1845-1899 \cite{LatamXIX}.

\subsection{Cleaning}

The cleaning step is essential for The British Library and Project Gutenberg datasets since some texts from these sources consisted solely of chapter, book, or newspaper titles, or were filled with numbers and other characters that added noise to the dataset. In the case of the LatamXIX dataset, these noisy rows were already filtered and complemented with an LLM OCR correction process that corrected many OCR errors within the corpus, making it cleaner and more fittable for the SSD task, as it preserves better semantic meaning for words and less noise. 

For The British Library books, an initial filter was applied using word confidence information to retain only those books with a mean OCR word confidence higher than 0.5. This experimental threshold was set to balance data loss ($2.26\%$ of rows) and text quality. After conducting several revisions with different examples, it was observed that this threshold maintained a high standard of text quality. Therefore, it was selected as the optimal balance between data retention and textual accuracy.

Same as in \citet{LatamXIX}, the cleaning steps to perform were:

\begin{enumerate}
    \item Remove duplicates and empty rows within the whole dataset. $6.94\%$ of rows were removed.
    \item Filter out rows where over 50\% of the characters are non-alphabetic, including spaces. $0.92\%$ of rows were removed.
    \item Remove the rows with 4 or fewer tokens. Samely, a new tokenizer was trained with a vocabulary size of 52,000, trained from the BETO pre-trained tokenizer \cite{CaneteCFP2020}. $0.50\%$ of rows were removed.
\end{enumerate}

These filters were applied to minimize the risk of compromising the results due to noise in the dataset.

\begin{figure}[t]
\centering
  \includegraphics[width=\columnwidth]{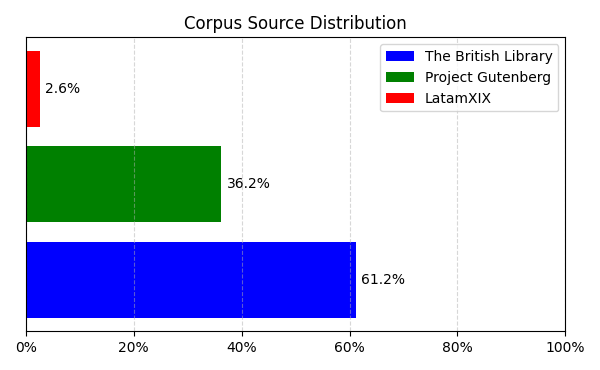}  \caption{Final corpus distribution by source. The percentage is computed over the total number of rows of the whole $C_{old}$ chunked corpus}
  \label{fig:data-chunked-1}
\end{figure}

\begin{table}[t]
    \centering
    \begin{tabular}{|c|c|}
    \hline
    \textbf{Feature} & \textbf{Value}  \\ \hline
    Size             & $\sim865 MB$ \\
    Rows             & $1,141,490$      \\
    Words           & $\sim125 M $ \\
    Tokens           & $\sim160 M$ \\
    Years Range       & 1800 - 1914    \\ \hline
    \end{tabular}
    \caption{\label{tab:data-chunked} Final $C_{old}$ chunked corpus information}
\end{table}

\subsection{Chunking}

As the historical texts from the corpus come from books and newspapers, many are very large, or some are very short with an average of $\sim110$ words and $\sim140$ tokens per text. For BERT-like models, the maximum sequence length consists of 512 tokens, which is insufficient for large texts like the current corpus texts.

Because of this, it's necessary to chunk the large texts within the dataset in a number shorter than 512 tokens. A much lower number was selected to make the chunked corpus fit for many different Language Models (LMs), for instance, a maximum of 256 tokens per text chunk, where a token was measured by training a new tokenizer over the cleaned version of the corpus\footnote{The final corpus can be found at \url{https://huggingface.co/datasets/Flaglab/spanish-corpus-xix} in all its three versions: "original", "cleaned", and "chunked"}.

\begin{figure}[t]
\centering  \includegraphics[width=\columnwidth]{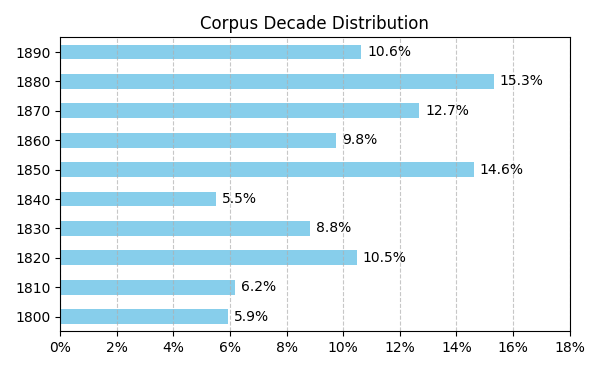}
  \caption{Final corpus distribution by decade. The percentage is computed over the total number of rows of the whole $C_{old}$ chunked corpus}
  \label{fig:data-chunked-2}
\end{figure}

During this step, over $67.6\%$ of the rows were chunked, adding 460,543 new rows. Each row was transformed into a part of a paragraph or left as a whole paragraph (no chunking) with no more than 256 tokens while preserving as much semantic meaning as possible. The preservation of semantic meaning in the chunked segments was achieved by splitting through punctuation marks and common paragraph-sentence separators. The rows distribution and corpus information can be found in Figures \ref{fig:data-chunked-1}, \ref{fig:data-chunked-2}, and Table \ref{tab:data-chunked} respectively. Also, the information on the Latin-American portion of the corpus can be found in Table \ref{tab:data-chunked-latam}.

\begin{table}[t]
    \centering
    \begin{tabular}{|c|c|}
    \hline
    \textbf{Feature} & \textbf{Value}  \\ \hline
    Size             & $\sim27 MB$ \\
    Rows             & $29.972$      \\
    Words           & $\sim4.5 M $ \\
    Tokens           & $\sim5.7 M$ \\
    Years Range       & 1845 - 1899    \\ \hline
    \end{tabular}
    \caption{\label{tab:data-chunked-latam} Final $C_{old}$ Latin-American portion chunked corpus information}
\end{table}

\section{Methodology}

\begin{figure*}[t]
\centering
  \includegraphics[width=\linewidth]{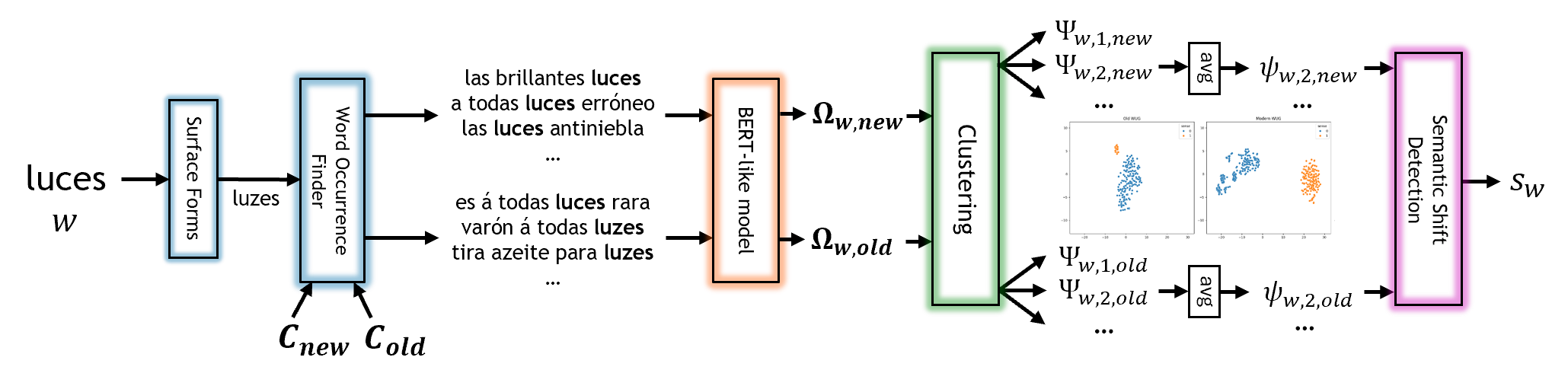}
  \caption{Historical Ink SSD Pipeline Architecture}
  \label{fig:pipeline}
\end{figure*}

To achieve effectively the desired task, and be able to perform a quality analysis of the results, we have defined the pipeline observed in Figure \ref{fig:pipeline}, with the following steps:

\begin{enumerate}
    \item Find the occurrences of a given word $w$ in $C_{old}$ and $C_{new}$ corpora.
    \item Retrieve the word embeddings in the found occurrences, using a BERT-like language model.
    \item Cluster the word usage by its meaning (sense), and average to get the centroids of the clusters.
    \item Perform the SSD task to identify lost/gained senses and measure the semantic change of the word ($s_w$).
\end{enumerate}

It's important to note that the pipeline was designed as a flexible and reusable solution for various contexts and configurable stages. Beyond analyzing the specific case of 19th-century Spanish, we propose a modular, plug-and-play pipeline with numerous adjustable stages. Each component of the pipeline can be used independently and configured for different use cases, ensuring versatility and adaptability for further research or applications.

\subsection{Find the Occurrences}

Given corpora $C_{old}$ and $C_{new}$, finding all texts where a word $w$ is used is straightforward when looking for \textit{exact occurrences}. However, this task becomes more complex with inflectional variations typical of languages like Spanish. For example, the word "crear" (to create) may appear as "creaste" (you created) or "creado" (created). Stemming can help by extracting the base form of the word, but it may lose some contextual meanings.

Also, in old Spanish, language rules have changed significantly, as noted by \citet{54f6385b-261f-3696-acfc-99604ae83a87}. These changes are detectable using the semi-automated framework presented as part of the Historical Ink project \cite{LatamXIX}, which extracts useful lists of \textit{surface forms} (i.e. specific appearance of a word in a given context) for words that underwent orthographic changes in 19th-century Latin-American Spanish (e.g., "luces" historically written as "luzes").

To address these challenges, we propose a method to find occurrences of a given word $w$ in diachronic corpora $C_{old}$ and $C_{new}$. This method organizes all word's expected usages and tokenizes both the word and the searching text, searching for each subword within a list of different orthographic forms of writing a given word.

For example, the word "gente" would be searched in $C_{old}$ as "gente", "jente" (surface form), "gent", or "jent" in that order. This method relies heavily on the tokenizer, so using one trained in the specific language is recommended for better performance.

\begin{figure*}[t]
\centering
  \includegraphics[width=\linewidth]{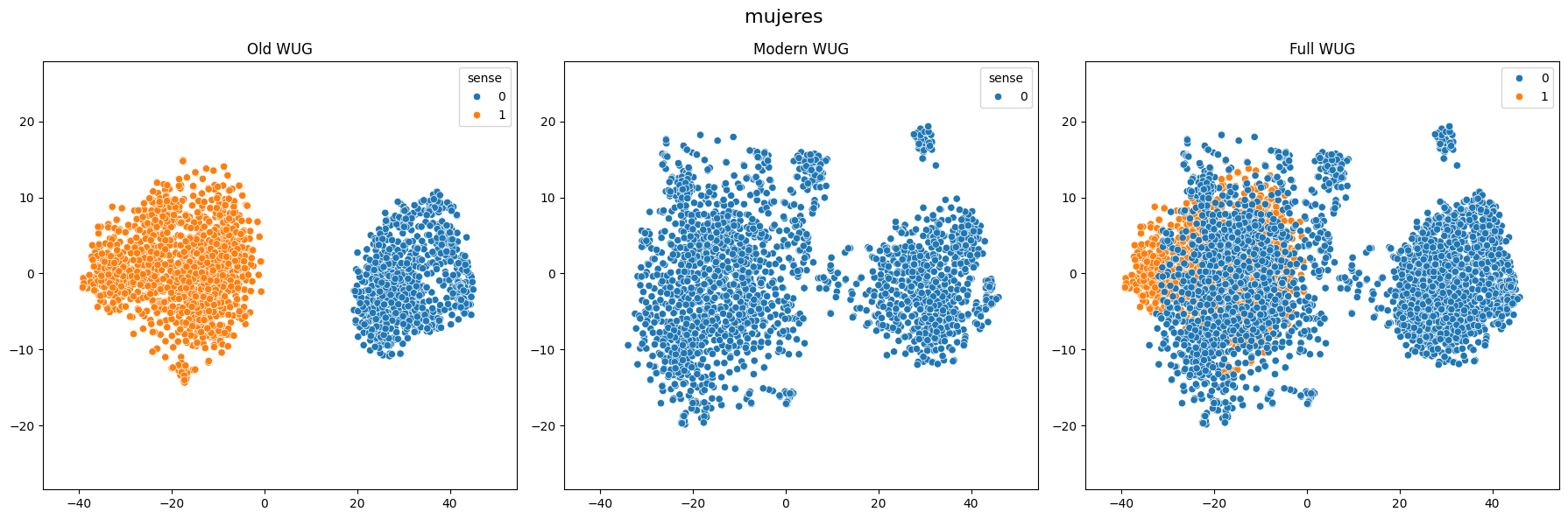}
  \caption{DWUG of the word "mujeres" (women), using the whole corpus fine-tuned model embeddings, the T-SNE dimensionality reduction algorithm, and the KMeans clustering algorithm (with the silhouette metric). Each color represents a meaning (cluster) of the word. The color changes between the left (old corpus) and center (modern corpus) images illustrate the overall semantic change between the two diachronic corpora.}
  \label{fig:clustering-sample-mujeres}
\end{figure*}

\subsection{Word Embeddings}

For the SSD task, contextual embeddings are very useful as they can capture the evolving meaning of words over time. By considering the surrounding context of a word within a sentence or document, contextual embeddings can provide an enhanced representation of its semantics, enabling the detection of particular shifts in meaning. In particular, there are some BERT-like LMs trained on Spanish corpora. Some of the most representative are BETO: Spanish Bert\footnote{ Available at \url{https://huggingface.co/dccuchile/bert-base-spanish-wwm-cased}} in both uncased and cased versions \cite{CaneteCFP2020}, Multilingual BERT\footnote{Available at  \url{https://huggingface.co/google-bert/bert-base-multilingual-cased}} in both uncased and cased versions, which has an important portion of training in Spanish \cite{BERT}, and AlBERT Spanish version\footnote{Available at 
 \url{https://huggingface.co/dccuchile/albert-base-spanish}}. All these models are BERT-based and have the same maximum sequence length of 512 tokens, BERT has an embedding size of 768, while ALBERT has a more compact embedding size of 128.

For this paper, we performed the SSD task using the mentioned LMs. Some were trained with the whole 19th-century Spanish corpus, while others were trained only with the Latin-American portion of the dataset. We fine-tuned these models using the specific corpus for each case, employing the Masked Language Modeling (MLM) task. In this task, 15\% of the corpus tokens were randomly masked, and the model learned to predict the masked tokens based on their context. This approach ensured that the model learned the unique linguistic style of each corpus, enabling it to generate word embeddings that accurately reflect the corpus' linguistic patterns, which is essential for detecting semantic shifts.

During the training phase, an Adam optimizer with a learning rate of $2\times 10^{-5}$ was employed, and the training proceeded with a batch size of 32, during a total of 5 epochs. Due to the low number of epochs, no Early Stopping was required, and the chosen parameters led to good resource utilization. The training time with the given batch size depended on the model but was on average 47 hours for the whole $C_{old}$ corpus, and 1 hour and 20 minutes for the Latin-American portion. The training was performed on an A40 GPU.

\subsection{Clustering}

We applied a joint clustering approach, combining both corpora within the same set of embeddings before clustering. Given two corpus $C_{old}$ and $C_{new}$, and a particular word $w$, the sets $\Omega_{w,old}$ and $\Omega_{w,new}$ are defined as the set of word embeddings generated in each corpus respectively, for the word $w$.

The clustering algorithm is meant to find the different meanings of a word within a given period, and overall the whole timespan of both $C_{old}$ and $C_{new}$ periods. This generates a well-known Diachronic Word Usage Graph (DWUG) for the word in both periods \cite{DWUG}, allowing to perform the semantic shift detection and change measurement between $old$ and $new$ periods, as seen in Figure \ref{fig:clustering-sample-mujeres}, where each color refers to a word meaning. 

The particular algorithms used were Affinity Propagation and KMeans with an automatic K finder under a certain score function such as silhouette score or inertia. The main problem with KMeans are words with a single meaning across the whole timespan. As common KMeans K-evaluation metrics are not fittable for one-cluster evaluation, so it wouldn't be possible to validate if the best number of clusters should be just one. As this occurs for many of the target words selected for analysis, a very good alternative for it is the Affinity Propagation (AP) clustering algorithm with a damping parameter of 0.975; this parameter was selected through a test with different values and a manually-driven evaluation of the number of clusters automatically selected by the algorithm. Selecting a high damping value for the AP algorithm leads to a more stable selection of the number of clusters as the requirements for new cluster creation are more strict, which is expected for this case.

The T-SNE dimensionality reduction algorithm was used to plot the DWUGs shown in this paper, with a perplexity of 50, which proved the best for better cluster space separation. For words with a lower number of found occurrences in the dataset, a lower perplexity was employed for its representation.

\subsection{Semantic Shift Measurement}

\begin{table*}[t]
\begin{tabular}{|cl|ccccccc|}
\multicolumn{2}{|c|}{LM}                                      & \multicolumn{2}{c|}{Average}                               & \multicolumn{3}{c|}{Clustering}                                                                                                                                                    & \multicolumn{2}{c|}{Non-Clustering}       \\ \hline
\multicolumn{1}{|c|}{\textbf{\#}} & Name                      & \multicolumn{1}{c|}{Clustering} & \multicolumn{1}{c|}{All} & \multicolumn{1}{c|}{AP} & \multicolumn{1}{c|}{\begin{tabular}[c]{@{}c@{}}KM\\ inertia\end{tabular}} & \multicolumn{1}{c|}{\begin{tabular}[c]{@{}c@{}}KM\\ silhouette\end{tabular}} & \multicolumn{1}{c|}{CD} & PRT             \\ \hline
\multicolumn{1}{|c|}{1}           & BETO cased FT             & 0.5799                          & 0.6017                   & \textbf{0.6124}         & 0.5598                                                                    & 0.5676                                                                       & 0.6285                  & 0.6402          \\
\multicolumn{1}{|c|}{2}           & \textbf{BETO cased LFT}       & \textbf{\textcolor{red}{0.5872 (1)}}             & 0.6064                   & 0.5853                  & 0.5815                                                                    & 0.5947                                                                       & 0.6307                  & 0.6396          \\
\multicolumn{1}{|c|}{3}           & BETO cased                & 0.5832                          & 0.6041                   & 0.5600                    & 0.5790                                                                     & \textbf{0.6107}                                                              & 0.6302                  & 0.6405          \\
\multicolumn{1}{|c|}{4}           & BETO uncased FT           & 0.5578                          & 0.5837                   & 0.5442                  & 0.5579                                                                    & 0.5714                                                                       & 0.6224                  & 0.6227          \\
\multicolumn{1}{|c|}{5}           & BETO uncased LFT     & 0.5658                          & 0.5890                    & 0.5594                  & 0.5676                                                                    & 0.5703                                                                       & 0.6223                  & 0.6255          \\
\multicolumn{1}{|c|}{6}           & BETO uncased              & \textbf{0.5862 (3)}             & 0.6043                   & 0.5916                  & 0.5819                                                                    & 0.5850                                                                        & 0.6167                  & 0.6463          \\
\multicolumn{1}{|c|}{7}           & mBERT cased LFT      & 0.5806                          & 0.5951                   & 0.5692                  & 0.5788                                                                    & 0.5939                                                                       & 0.6163                  & 0.6172          \\
\multicolumn{1}{|c|}{8}           & mBERT cased               & 0.5782                          & 0.5949                   & 0.5675                  & 0.5808                                                                    & 0.5863                                                                       & 0.6100                    & 0.6297          \\
\multicolumn{1}{|c|}{9}           & mBERT uncased LFT    & 0.5593                          & 0.5929                   & 0.553                   & 0.5633                                                                    & 0.5615                                                                       & 0.6405                  & 0.6464          \\
\multicolumn{1}{|c|}{10}          & mBERT uncased             & 0.5762                          & 0.6065                   & 0.5523                  & 0.5924                                                                    & 0.5839                                                                       & \textbf{0.6457}         & 0.6581          \\
\multicolumn{1}{|c|}{11}          & AlBERT LFT & 0.5717                          & 0.5928                   & 0.5731                  & 0.5796                                                                    & 0.5624                                                                       & 0.6160                   & 0.6328          \\
\multicolumn{1}{|c|}{12}          & AlBERT          & \textbf{0.5869 (2)}             & \textbf{0.6132}          & 0.5758                  & \textbf{0.5992}                                                           & 0.5857                                                                       & 0.6373                  & \textbf{0.6682}
\end{tabular}
    \caption{\label{tab:benchmark} LM benchmark through the LSCDiscovery \cite{LSCD} F1 of the Binary Change Detection task. Each model was fine-tuned for the Latin-American corpus (LFT) and both BETO-cased and uncased models were also fine-tuned for the whole corpus (FT), comparing also with non-fine-tuned versions.}
\end{table*}

Once clustering is performed, the measurement for Semantic Shift is straightforward. There are two main divisions of the SSD task which are Binary Change Detection (BCD) and Graded Change Detection (GCD) \cite{LSCD}, where Graded Change Detection is the most common and useful, but also the most challenging task for change classification, which consists of ranking a list of target words based on their degree of change \cite{periti2024systematic}.

The consolidation of techniques for measuring semantic shift detection has been a high-growth area, with the proposal of many different techniques, some of them comparing sets of embeddings (e.g. the clusters), and others comparing individual embeddings (e.g. the centroids). \citet{SSD} present a survey that compiles many of the most used state-of-the-art techniques for grading the semantic shift of a word between two temporal-different corpora, classifying them between form- and sense-based approaches. 

Given $m$ number of clusters (senses) for the word $w$, returned by the clustering algorithm, we define $\Psi_{w, s, t}$ as the cluster with the sense $s$ for the word $w$ in the period $t$, such that all the senses compound the whole set of embeddings.

\begin{equation}
    \Omega_{w,t} = \bigcup_{s=1}^{m} \Psi_{w, s,t}\quad \forall t = \{new, old\}
\end{equation}

For these clusters, a centroid embedding is computed as the average:

\begin{equation}
    \psi_{w, s, t} = \hbox{avg}(\Psi_{w, s, t})  \quad \forall t = \{new, old\}
\end{equation}

Finally, two different formulas were taken from \citet{SSD} to measure the semantic shift $f$, based on the cosine similarity function (CS). With this shift, for each word, we would have as many semantic shifts $f$ as the number of clusters given by the algorithm ($m$), so we could determine which senses have had a diachronic shift and which haven't, for each word.\\

\textbf{Cosine Distance (CD):}

\begin{equation}
    f_{CD}(w,s) = 1 - CS(\psi_{w, s, old}, \psi_{w, s, new}) 
\end{equation}

\textbf{Inverted similarity over Word Prototype (PRT):}
\begin{equation}
    f_{PRT}(w,s) = \frac1{CS(\psi_{w, s, old}, \psi_{w, s, new})} 
\end{equation}

It should be noted that if a sense is not present within a period, whether old or new period, $f_{CD}$ should be 1.0, meaning a complete change of the given sense. If the sense is absent from the embeddings of the old period ($\Psi_{w,s,old} = \emptyset$), it means that the sense was gained in modern Spanish; otherwise, if the sense only exists in the embeddings of the old period ($\Psi_{w,s,new} = \emptyset$), it means that the sense was lost in modern Spanish, as seen in Figure \ref{fig:clustering-sample-mujeres} where the sense 1 (orange color) is not present in the modern WUG.

For this task, it is crucial to consider the frequency of points per cluster within each period. If a cluster has significantly fewer points in a period, specifically less than 10\% of the total, we classify these points as either misclassifications or obsolete words. This allows us to treat the cluster as a gained or lost sense. We chose this threshold based on testing with few known examples, where it provided the best performance in detecting gained and lost senses.

\section{Evaluation and Model Selection}

As mentioned, several pre-trained Language Models (LMs) are available for large Spanish corpora. We needed an evaluation method to select the best model for our analysis. The LSCDiscovery shared task \cite{LSCD} provides over 65,000 annotated examples for 100 target words using the DURel framework proposed by \citet{DUREL}. This annotated corpus is highly useful for evaluating the LMs, as its time period is within the 19th century. Even though the LSCDiscovery task differs from the one in this paper, it offers a valuable benchmark. Our task focuses on detecting the different meanings of a word in a diachronic corpora and measuring their semantic shift over time. By comparing with the LSCDiscovery task, we ensure a rigorous evaluation, confirming that the models are robust and effective across various contexts and not overly tailored to a single specific task.

The task's corpus includes pairs of sentences rated from 1 to 4, where 1 indicates identical word usage and 4 indicates completely different usage \cite{DUREL}. To evaluate the models, we converted this numerical assessment into a binary evaluation: ratings 1-2 indicated no semantic change, while ratings 3-4 indicated a semantic change. We then defined five specific methods to classify a pair of word uses as either semantic change (1) or no change (0). Among these, two methods — \textit{cosine distance} (CD) and \textit{inverted similarity over word prototype} (PRT) — were tested purely for task purposes. However, the methods of primary importance for this paper are those related to sense clustering.

The three clustering-based evaluation methods consist of grouping all the embeddings of the occurrences of a word, as mentioned in the SSD section. Then, given two uses, if they do not belong to the same cluster, a semantic change is indicated (1); otherwise, no semantic change is indicated (0). This was evaluated using Affinity Propagation and KMeans (with silhouette and inertia metrics) methods. Finally, the model with the best average results across the three clustering methods was selected. The benchmark results can be seen in Table \ref{tab:benchmark}. 

While the results provide valuable insights into the models' capabilities, they should not be directly compared to those from the LSCDiscovery leaderboard \cite{LSCD}. Instead, they serve as an effective benchmark for assessing how well the LMs perform in detecting semantic changes within our specific historical context. The differences in tasks and the method approaches for our study reflect that direct comparisons with LSCDiscovery scores are not applicable.

Given the results, the best-performing model was BETO fine-tuned on the Latin American dataset\footnote{Fine-tuned model was uploaded to HuggingFace and is available at \url{https://huggingface.co/Flaglab/beto-cased-finetuned-xix-latam}}. A possible explanation for this is that the Latin American portion of the corpus underwent an additional step of LLM OCR correction, which removed OCR-related errors and produced cleaner text. This likely reduced noise and improved the quality of fine-tuning. Additionally, BETO was trained solely in Spanish, unlike multilingual BERT, which was trained in many different languages. According to \cite{CaneteCFP2020}, this single-language focus tends to result in better performance compared to multilingual models. This model was the one used for evaluating the target words and creating the DWUGs presented in appendix \ref{sec:appendix3}.

\section{Results}

The results of the trained model focus on a specific group of 255 target words\footnote{From all 255 words, only 233 had enough occurrences in the modern corpus. The DWUGs and SSD for both AP and KMeans algorithms are available in the notebook \url{https://colab.research.google.com/drive/1eaULQocxyuCNX0ftBvDJwe8nfpEi5s6i}} selected for their historical significance and relevance to generate hypotheses about potential semantic shifts over time, confirming the consistency of the results. Some examples of the DWUGs analyzed in this section are available in Appendix \ref{sec:appendix3} for both AP and KMeans.

One of the main results of this research was to highlight the success and failure cases for both AP and KMeans clustering algorithms, as both were used to compute the senses of all 255 words. Affinity Propagation (AP) performed poorly in many cases where it couldn't detect multiple usages of a word, such as "grave" (serious/bass), or detected many different senses for other words, such as "honor" (honour), as shown in Figure \ref{fig:clustering-sample-01}. However, it effectively detected single-sense words, a task that KMeans wasn't capable of due to metrics used to choose the best K. However, KMeans performed very well in most cases, effectively detecting and clustering the senses of multi-meaning words over time.

As displayed in Figure \ref{fig:clustering-sample-01}, some words like “rey” (king) and “usurer” (usurer) present neither polysemy nor notable historical changes. However, the term “mujeres” (women), as shown in Figure \ref{fig:clustering-sample-mujeres}, shows a change in modern usage. This finding is particularly interesting in the context of both historical discourse analysis in gender studies and historical linguistics studies, as it is an example of computational verification.

The semantic transformation of the word women, as plotted in Figure \ref{fig:clustering-sample-mujeres} and in Appendix \ref{sec:appendix2}, primarily pertains to the antiquated use of “mujeres” designating a particular group of female individuals. In 19th-century Spanish, lexical tradition mandated the rigorous use of masculine forms of nouns and adjectives as the universal form, encompassing both genders (feminine and masculine) \cite{Porto_Dapena_1975}. Thus, the word “hombres” (men) could be used as a synonym for humanity, while the use of “mujeres” (women) was more likely to be reserved for describing a private group of women. Twentieth-century gender studies introduced a unified meaning to the word “mujeres”. Joan W. Scott famously stated that "-Women's experience- or -women's culture- exists only as the expression of female particularity in contrast to male universality" \cite{Scott1988-im}. This idea explains the rupture in the modern usage of the word women towards the relational concept of gender in the 20th century \cite{Lux_Pérez_Pérez_2020}. 

Consequently, the term “mujeres” evolved from a specific designation to a broader and more inclusive reference, reflecting significant social and cultural shifts in gender discourse. As we have observed, the contemporary usage of “mujeres” tends to encompass all women more generically, since it was not until the 20th century that historical consideration began to differentiate "women" as a collective separate from "men". In the past, the term was used to refer to a distinct group of women, thereby distinguishing women from other plural nouns such as men, children, or even animals. Modern usage of "women" almost exclusively serves to differentiate women from men.

Other insightful results demonstrate both how the polysemy of words changes over time, as seen in examples in Appendix \ref{sec:appendix1}, and the particularities of word semantics diachronically used in Latin American Spanish. Historical linguistics studies acknowledge "El español de América" as a main Spanish variant, for which corpus studies are yet to be conducted. Newspapers are recognized as a legitimate source for exploring the particularities of linguistic variants \cite{Gutierrez_Mate2023-xk}. Hence, the LatamXIX dataset we used to model the quantitative experiments might initiate a triangulation with new regional research. For example, we have observed how the term “infancia” (infancy/childhood), as depicted in Figure \ref{fig:clustering-sample-02}, was predominantly used in the 19th century as an abstract reference to the nascent phase of objects, entities, or people. This suggests a metaphorical use of the word, indicative of a broader, symbolic interpretation of "infancy" or "early development" during this era.

Newly formed Latin American nations in the 19th century viewed themselves as children recently independent from their mother, metropolitan Spain. Consequently, the term "infancia de la patria" (infancy of the nation) described the contradictory and highly unstable political and social times experienced in Latin America during that era. These old meanings have largely been supplanted by the modern understanding of "childhood", which specifically refers to the population segment of children. These results align with the second wave of human rights in the 20th century, which expanded the 19th century's initial civil rights to include specific rights for various western population groups, such as children and women.

Words like "sentimiento" (sentiment) have lost one of their historical meanings, as illustrated in Figure \ref{fig:clustering-sample-02}. In contemporary usage, "sentimiento" serves as a synonym for the feelings experienced by an individual or group of people. However, one of its older meanings has almost disappeared. In the 19th century, "sentimiento" was used to describe the expression of a person's correctness, effectively acting as a synonym for morality, or even referring to someone's elevated religious or artistic spirit. On the other hand, the term "sublime" (sublime or elevated) has largely fallen out of use and is scarcely found in the modern dataset, as depicted in Figure \ref{fig:clustering-sample-02}. Appendix \ref{sec:appendix2} contains examples of the 255 words' semantic shift detection outputs, including other examples such as "luces" (ideas/lights) and "servidores" (servants/servers).

\begin{figure}[t]
\centering
  \includegraphics[width=\columnwidth]{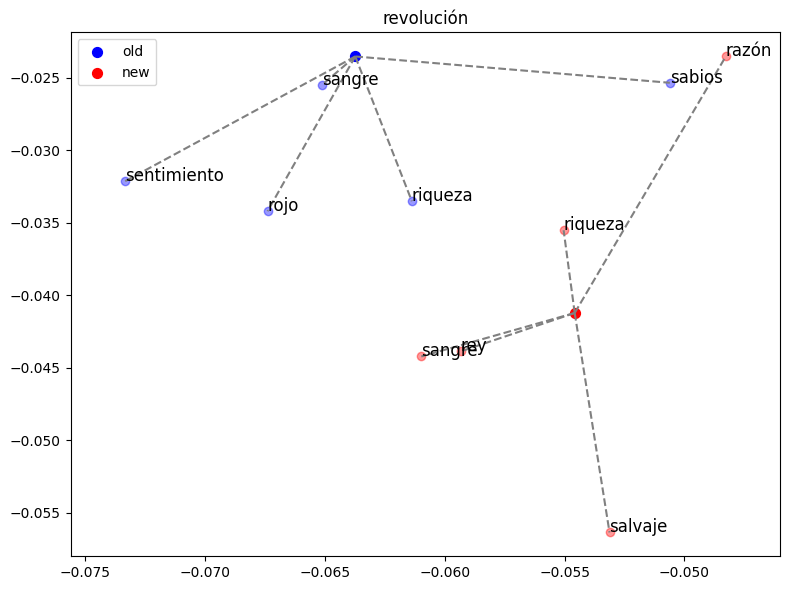}  
  \caption{Diachronic comparison of word "revolución" (revolution) and its related words, between the old and the modern period using PCA dimensionality reduction algorithm.}
  \label{fig:pca}
\end{figure}

Finally, word comparison also proves highly valuable for numerous diachronic analyses. In each period, the most representative sense of a word is determined based on its frequency dominance among other senses. Then, its sense cluster centroid is computed to allow comparison between words. Within the set of 255 words, the 5 words exhibiting the highest cosine similarity to this centroid are selected, indicating their related usage contexts. For example, as observed in Figure \ref{fig:pca}, the word "revolución" (revolution) historically exhibited close associations with the words blood, richness, feeling, wise, and red. In contemporary contexts, however, the term "revolución" is linked to terms like king, reason, and savage, and it remains related to blood and richness in different proportions, with blood now more distant and richness closer.


This study provides significant insights into the SSD of 19th-century Spanish words, utilizing computational linguistics to uncover shifts in word meanings relevant to both global and Latin American contexts. By developing a specialized corpus and employing methods such as fine-tuning BERT-like models and diachronic word embeddings, we achieved a nuanced analysis of historical semantic changes. Our examination of selected words reveals the relation between societal, cultural, and political events and the shift of words' semantic meaning over time.

The application of SSD and modern computational techniques highlights the evolution of linguistic analysis from manual to systematic approaches, enhancing the accuracy of semantic shift detection and deepening our understanding of language as a dynamic entity. This study's interdisciplinary implications are notable, offering potential benefits to fields like history, sociology, and digital humanities, where these insights can provide deeper context to historical cultural shifts.

Looking ahead, the methodologies and findings of this project can serve as a framework for future research in other languages and periods, suggesting a scalable approach to historical linguistics and semantic analysis. The flexible and reusable pipeline developed here can be adapted for various contexts and stages. Future research could apply this pipeline with modified parameters or data for different use cases or languages, to prove its performance on different contexts.

However, an evaluation of the selected models for the Latin-American corpus, particularly for clustering, is still needed. An annotated dataset similar to the given in the AXOLOTL-24 shared task \cite{Axolotl24}, but for Latin-American Spanish, would be highly beneficial. Such a dataset, with examples of specific word usages, their periods, and a gold standard for word senses, would enable a more focused assessment of the models beyond the task evaluation presented in Table \ref{tab:benchmark}.

\section{Acknowledgements}

We would like to thank the three anonymous reviewers from the ACL 2024 LChange'24 conference for their helpful feedback and suggestions.

\bibliography{acl_latex}

\appendix
\renewcommand\thefigure{\thesection\arabic{figure}}
\renewcommand\thetable{\thesection\arabic{table}}
\setcounter{figure}{0}    
\setcounter{table}{0}    

\section{Usage Examples per Sense}
\label{sec:appendix1}
\begin{figure*}[t]
\centering
    \includegraphics[width=\linewidth]{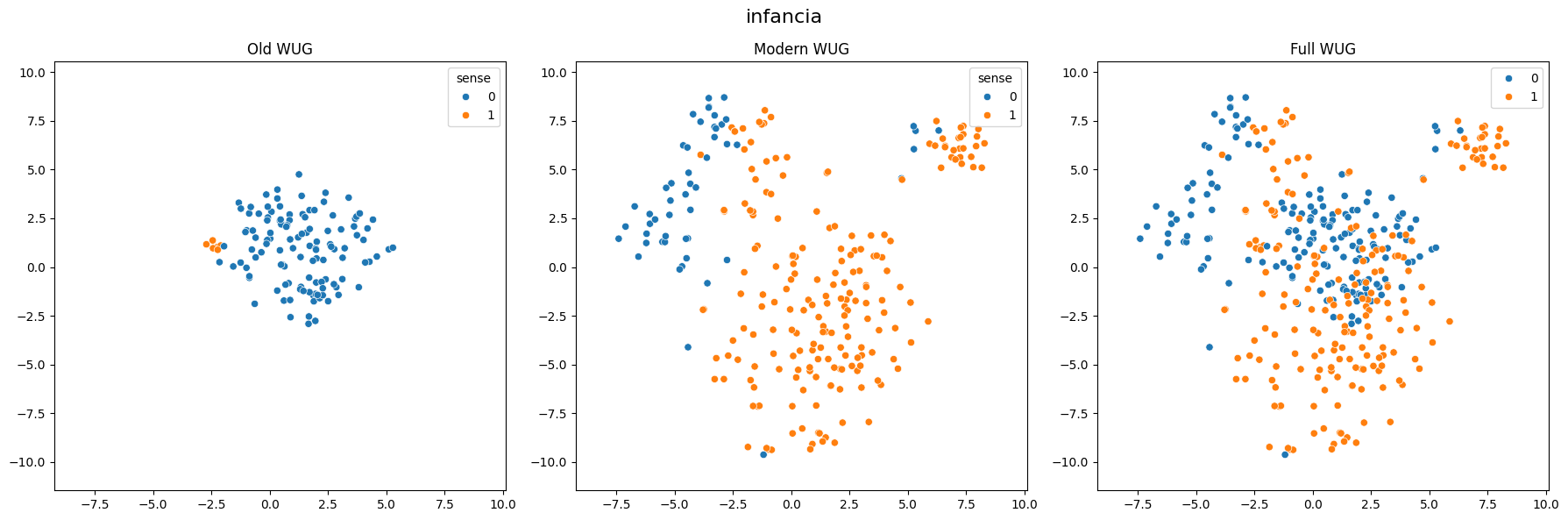}
    
    \includegraphics[width=\linewidth]{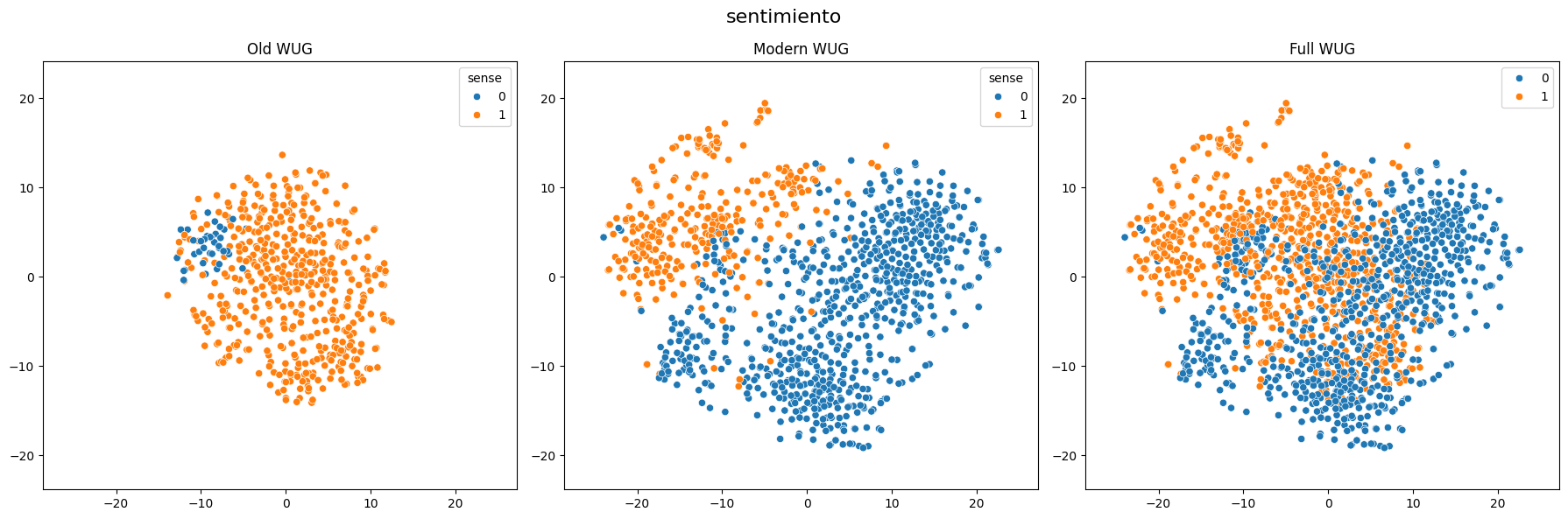}

    \includegraphics[width=\linewidth]{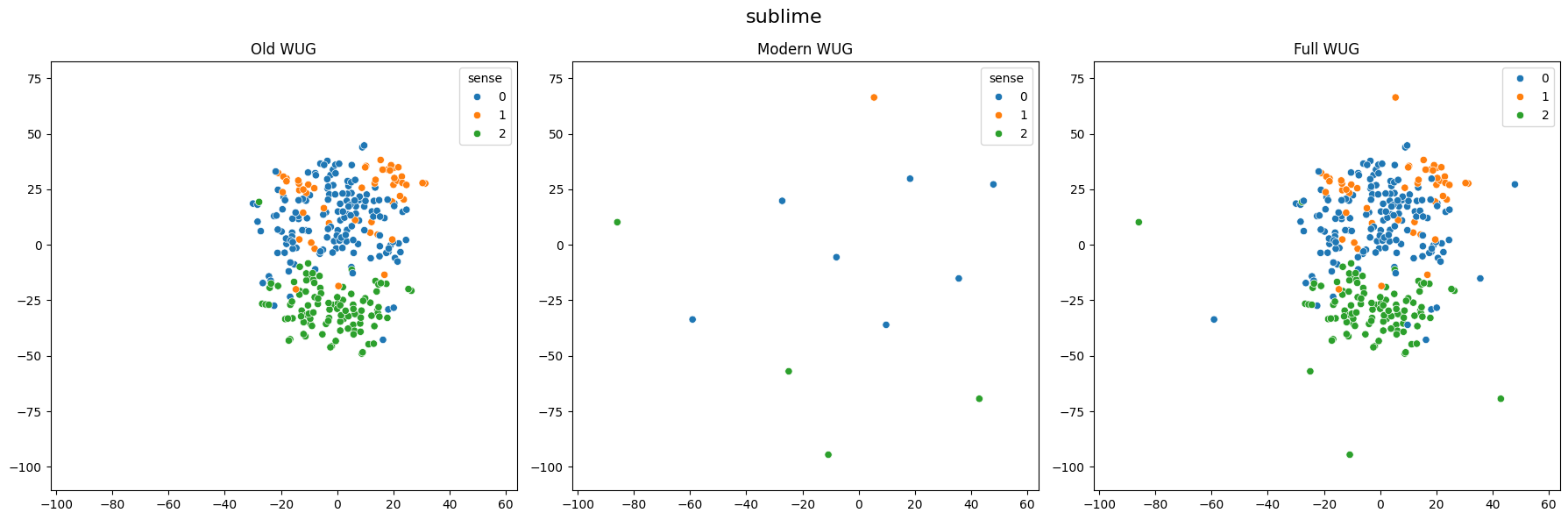}

    \includegraphics[width=\linewidth]{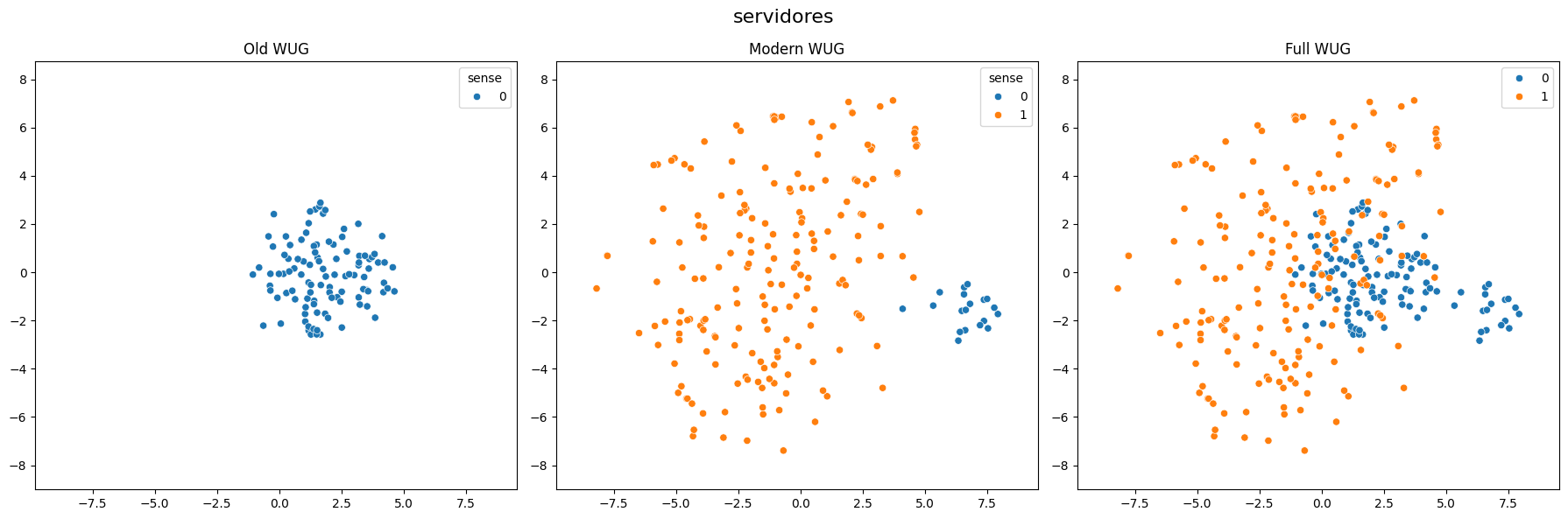}

    \caption{DWUG using the Latin American portion of the corpus fine-tuned model embeddings, the T-SNE dimensionality reduction algorithm, and the \textbf{KMeans} clustering algorithm (with the silhouette metric). All words are correctly clustered.}
  \label{fig:clustering-sample-02}
\end{figure*}

\textbf{Infancia:} The word has presented a semantic shift as shown in Figure \ref{fig:clustering-sample-02}

\textcolor{blue}{Sense 0 in New}-"Adopción derechos del niño, protección de la \textcolor{red}{infancia}, tráfico de personas". 

\textcolor{blue}{Sense 1 in Old}-"Los pueblos, como los hombres, tienen su \textcolor{red}{infancia}, embrion todavía entre nosotros, período delicado y peligroso, en el que todo exceso é indiscreción trastorna el organismo é impide el desarrollo, si es que no lo destruye." "Escamilla, por ejemplo, se casó desde la \textcolor{red}{infancia} con una matrona llamada Portería del Congreso de Escamilla: lleva dos apellidos, esta señora, no porque sea bigama, (pues no ha tenido mas que un solo marido) sino porque su papà es el señor Congreso, un viejo, mui necio."

\textbf{Sentimiento:} The word has presented a semantic shift as shown in Figure \ref{fig:clustering-sample-02}

\textcolor{blue}{Sense 0 in New}- "65\% de la personas que expresan un \textcolor{red}{sentimiento} personal de temor o esperanza". "Reforzar entre los europeos el \textcolor{red}{sentimiento} de pertenencia a una misma Comunidad". 

\textcolor{blue}{Sense 1 in Old}- "Será un gran artista de mucho \textcolor{red}{sentimiento}, posee una rica voz, si la educa, y tiene mucho aplomo en las tablas, es feo, pero simpático". "Una forma de expresión nueva, en la que brillaban un profundo \textcolor{red}{sentimiento} poético y una suerte de ingenuidad". "Que el divino arte de la música, lenguaje de la inteligencia y del \textcolor{red}{sentimiento}, ejerce sobre todos los hombres una influencia poderosa, que al mismo tiempo que atempera las pasiones, despierta las ideas de moralidad y de sociabilidad". "Republicano de ideas y de \textcolor{red}{sentimiento}, ha sabido armonizar sus opiniones políticas con sus creencias".

\textbf{Sublime:} The word presented polysemy in the past but is no longer in use as shown in Figure \ref{fig:clustering-sample-02}

\textcolor{blue}{Sense 0 in Old}-"Hé aquí un epitafio \textcolor{red}{sublime}; la madre que busca al hijo bajo la sombra de los laureles, en la soledad de la muerte como dos almas inseparables, siempre unidas, siempre amantes". 

\textcolor{blue}{Sense 1 in Old}- "Bolívar, el del genio \textcolor{red}{sublime} que todo lo abarcó, que todo lo comprendió, y á quien debieron su existencia y su gloria, en menos de un cuarto de siglo, la mayor parte de las nacionalidades del Nuevo Mundo".

\textcolor{blue}{Sense 2 in Old}- "á veces las leyes naturales-puede sí ejercer el \textcolor{red}{sublime} ministerio de aliviar (obra divina, según Hipócrates) y consolar á los que sufren."

\textbf{Servidores:} The word gained a new sense as shown in Figure \ref{fig:clustering-sample-02}

\textcolor{blue}{Sense 0 in Old}-"Era allí donde se alojaba el Cacique, su familia y sus principales \textcolor{red}{servidores}". "a depositar- sus votos en favor de los buenos y leales \textcolor{red}{servidores} de la causa".

\textcolor{blue}{Sense 0 in New}- "si la joven no está en un convento, rodearla de \textcolor{red}{servidores} que la acompañen por todas partes". "La Comisión y nosotros somos los \textcolor{red}{servidores} de los ciudadanos de nuestros Estados miembros".

\textcolor{blue}{Sense 1 in New}- "la adquisición o el alquiler de ordenadores personales, \textcolor{red}{servidores} y microordenadores". "operación de los sistemas y de la red, y \textcolor{red}{servidores} para bases de datos, la Web, el FTP".

\section{SSD Examples}
\label{sec:appendix2}

Some of the SSD results chosen were selected from Affinity Propagation algorithm clusterization, particularly those with only one sense such as "rey" and "usurero".

\begin{table}[h]
\begin{tabular}{c|cc|ccc|}
\textbf{}                                                                                      & \multicolumn{1}{c|}{Word} & Sense      & \multicolumn{1}{c|}{CD} & \multicolumn{1}{c|}{PRT} & \begin{tabular}[c]{@{}c@{}}gained/lost\\ Sense\end{tabular} \\ \hline
\multicolumn{1}{|c|}{\multirow{2}{*}{\rotatebox[origin=c]{90}{AP}}}                                                                         & Rey                       & 0          & 0.005                   & 1.005                    &                                                             \\ \cline{2-6}
\multicolumn{1}{|c|}{}                                                                         & Usurero                     & 0          & 1.0                   & $\infty$                    & \textbf{lost}                                                             \\ \hline 
\multicolumn{1}{|c|}{\multirow{11}{*}{\rotatebox[origin=c]{90}{KMeans}}}     & Luces                     & 0          & 0.012                   & 1.012                    &                                                             \\
\multicolumn{1}{|c|}{}                                                                         & Luces                     & 1          & 0.012                   & 1.013                    &                                                             \\ \cline{2-6} 
\multicolumn{1}{|c|}{} & Infancia                  & 0          & 0.017                   & 1.017                    &                                                             \\
\multicolumn{1}{|c|}{}                                                                         & Infancia                  & \textbf{1} & 1.0                     & $\infty$                 & \textbf{gained}                                             \\ \cline{2-6} 
\multicolumn{1}{|c|}{}                                                                         & Sentimiento               & \textbf{0} & 1.0                       & $\infty$                 & \textbf{gained}                                             \\
\multicolumn{1}{|c|}{}                                                                         & Sentimiento               & 1          & 0.003                   & 1.003                    &                                                             \\ \cline{2-6} 
\multicolumn{1}{|c|}{}                                                                         & Sublime                   & \textbf{0} & 1.0                     & $\infty$                 & \textbf{lost}                                               \\
\multicolumn{1}{|c|}{}                                                                         & Sublime                   & \textbf{1} & 1.0                     & $\infty$                 & \textbf{lost}                                               \\
\multicolumn{1}{|c|}{}                                                                         & Sublime                   & \textbf{2} & 1.0                     & $\infty$                 & \textbf{lost}                                               \\ \cline{2-6} 
\multicolumn{1}{|c|}{}                                                                         & Servidores                & 0          & 0.043                   & 1.045                    &                                                             \\
\multicolumn{1}{|c|}{}                                                                         & Servidores                & \textbf{1} & 1.0                     & $\infty$                 & \textbf{gained}                                            
\end{tabular}
    \caption{\label{tab:ssd} SSD for some of the 255 target words; the ones mentioned in the paper, and others added in the appendix DWUGs.}
\end{table}

\section{DWUGs Examples}
\label{sec:appendix3}

\newpage

\begin{figure*}[t]
\centering
    \includegraphics[width=\linewidth]{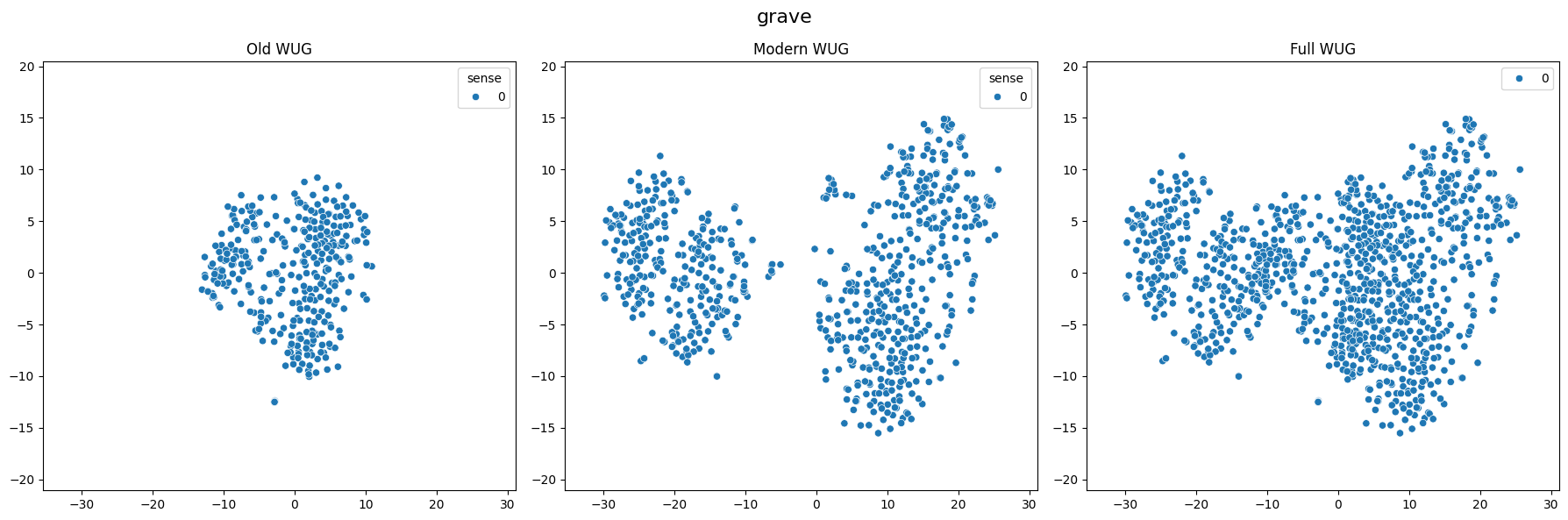}

    \includegraphics[width=\linewidth]{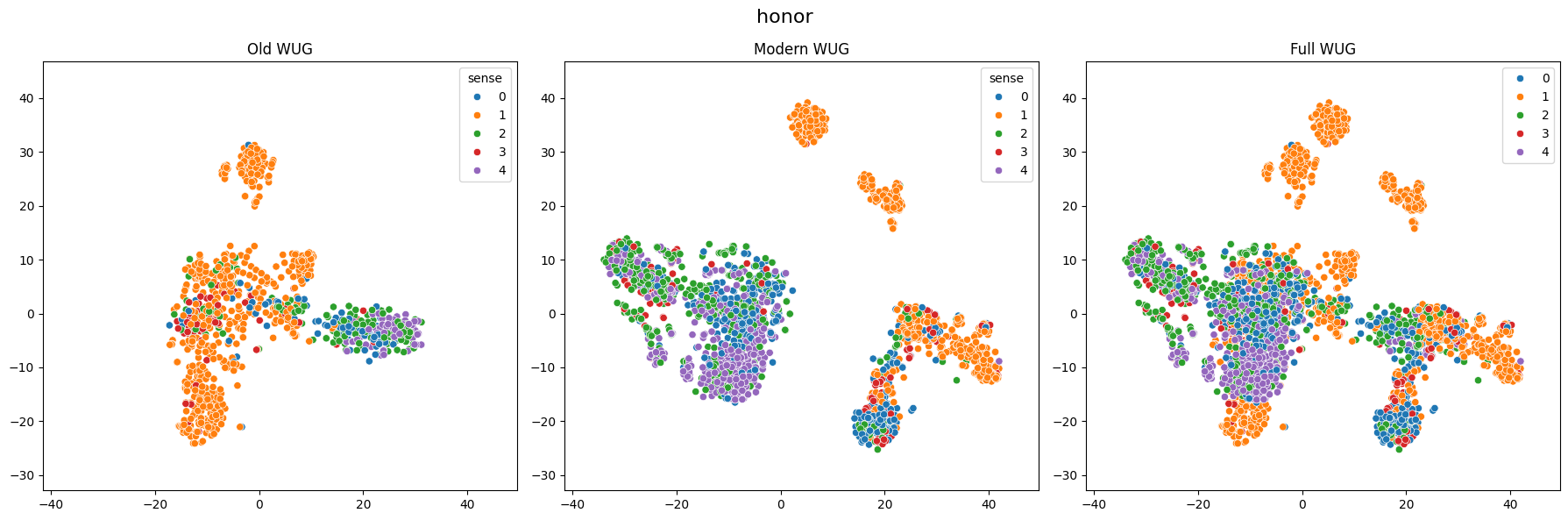}

    \includegraphics[width=\linewidth]{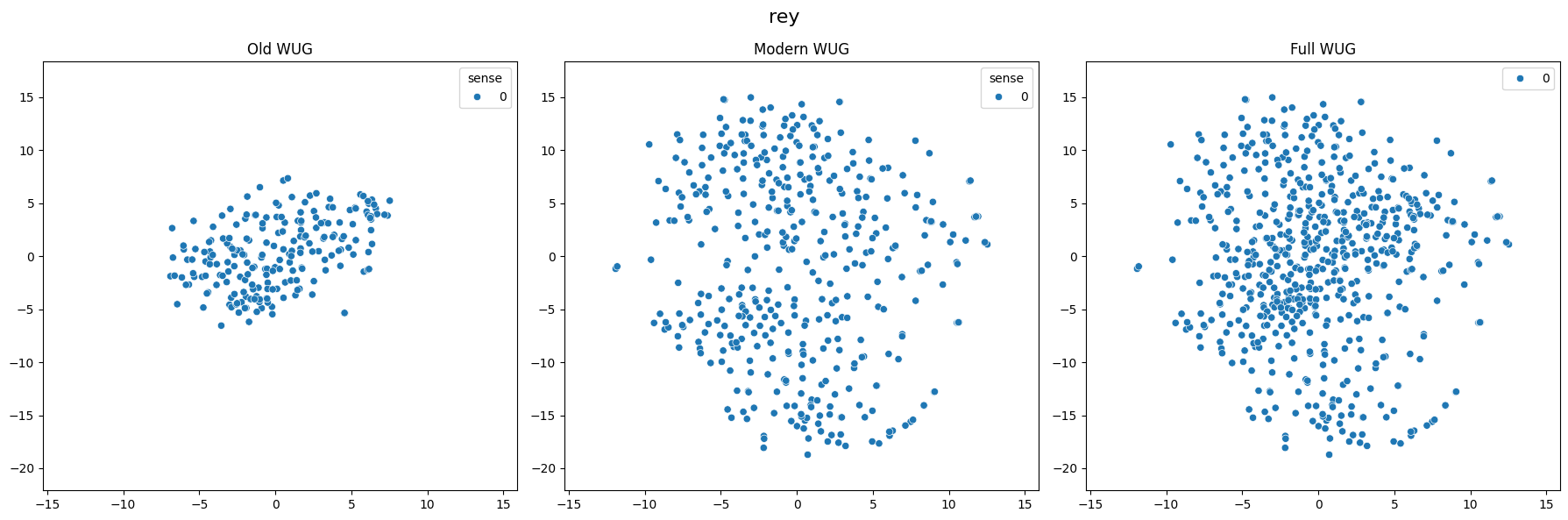}

    \includegraphics[width=\linewidth]{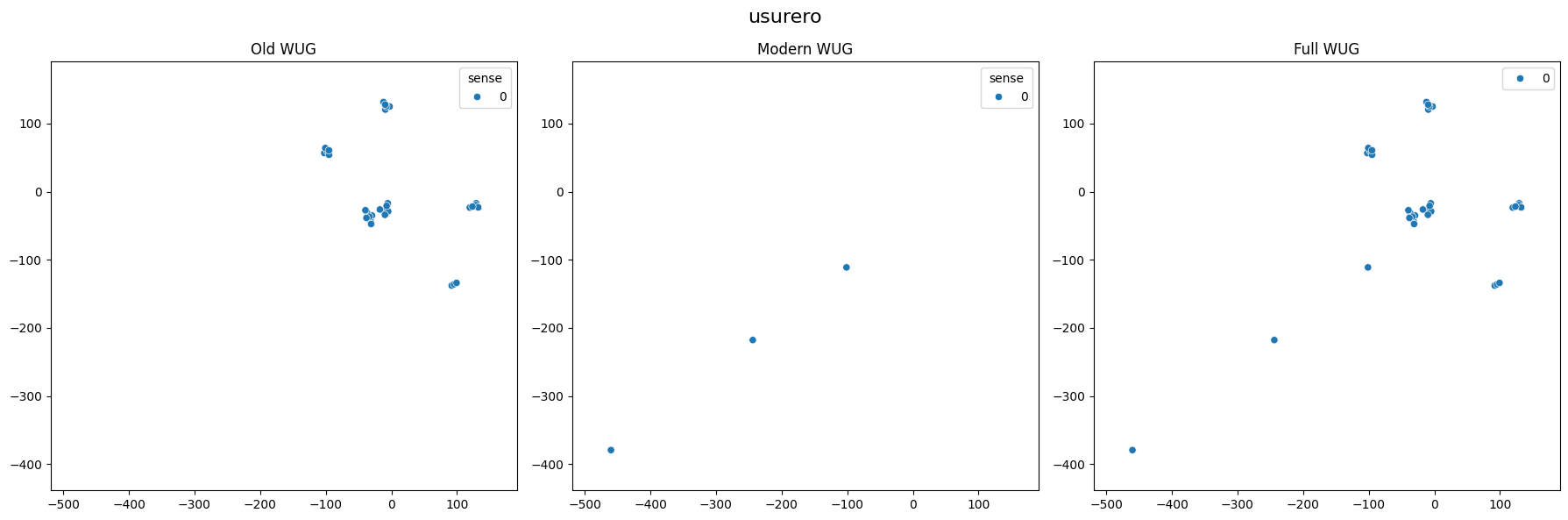}

    \caption{DWUG using the Latin American portion of the corpus fine-tuned model embeddings, the T-SNE dimensionality reduction algorithm, and the \textbf{Affinity Propagation} clustering algorithm. Words "grave" and "honor" are wrong clustered, and words "rey" and "usurero" are correctly clustered.}
  \label{fig:clustering-sample-01}
\end{figure*}

\end{document}